\documentclass{llncs}
\usepackage{array}
\usepackage{graphicx}
\usepackage{booktabs}
\usepackage{siunitx}
\usepackage{arabtex}
\usepackage{utf8}
\usepackage{makecell}

\setcode{utf8}

\begin{document}
\title{A Combined CNN and LSTM Model for Arabic Sentiment Analysis}
\author
{Abdulaziz M. Alayba\inst{1} \and
Vasile Palade\inst{2} \and
Matthew England\inst{3}\and
Rahat Iqbal\inst{4}
}
%
%\authorrunning{A. Alayba et al.}
\institute{School of Computing, Electronics and Mathematics
	\\
	Faculty of Engineering, Environment and Computing,
	Coventry University, UK \\
	\email{Alaybaa@uni.coventry.ac.uk\inst{1} \\
	\{Vasile.Palade\inst{2}, Matthew.England\inst{3}, R.Iqbal\inst{4}\}@coventry.ac.uk
}}
\maketitle              % typeset the header of the contribution
\begin{abstract}
	Deep neural networks have shown good data modelling capabilities when dealing with challenging and large datasets from a wide range of application areas. Convolutional Neural Networks (CNNs) offer advantages in selecting good features and Long Short-Term Memory (LSTM) networks have proven good abilities of learning sequential data. Both approaches have been reported to provide improved results in areas such image processing, voice recognition, language translation and other Natural Language Processing (NLP) tasks. Sentiment classification for short text messages from Twitter is a challenging task, and the complexity increases for Arabic language sentiment classification tasks because Arabic  is a  rich language in morphology. In addition, the availability of accurate pre-processing tools for Arabic is another current limitation, along with limited research available in this area. In this paper, we investigate the benefits of integrating CNNs and LSTMs and report obtained improved accuracy for Arabic sentiment analysis on different datasets. Additionally, we seek to consider the morphological diversity of particular  Arabic words by using different sentiment classification levels.

\end{abstract}
\section{1.	Introduction}
In the past decade, social media networks have become a valuable resource for data of different types, such as texts, photos, videos, voices, GPS reading, etc. The explosion of data we experience today in many areas has led researchers in data science to develop new machine learning approaches. There were improvements in different areas, such as: Neural Networks, Deep Learning, Natural Language Processing (NLP), Computer Vision, Geolocation Detection, etc.  Sentiment Analysis is one of the topics that attracted much attention from NLP and machine learning researchers. Sentiment analysis deals with the texts or the reviews of people that include opinions, sentiments, attitudes, emotions, statements about products, services, foods, films, etc.~\cite{Liu2012}. 

There is a certain sequence of steps to perform supervised learning for sentiment analysis, i.e., converting the text to numeric data and mapping with labels, performing feature extraction/selection to train some classifiers using a training dataset and then estimate the error on the test dataset. Sentiment analysis has various analytic levels that are: document level, sentence level, aspect level~\cite{Balaji2017}~\cite{Feldman2013}, word level, character level~\cite{Lakomkin2017} and sub-word level~\cite{Joshi2016}. Deep neural networks have shown good performance in this area in~\cite{Kim2014},~\cite{Yin2015} and~\cite{Shin2017}.

We have also obtained good results  on using deep neural networks for sentiment analysis on our own dataset, an Arabic Health Services dataset, reported in~\cite{Alayba2017} and~\cite{Alayba2018}. We have obtained an accuracy between 0.85 and 0.91 for the main dataset in~\cite{Alayba2017} using SVM, Na\"{\i}ve Bayes, Logistic Regression and CNNs. Also, using merged lexicon with CNNs and pre-trained Arabic word embedding, the accuracy for the main dataset was improved to 0.92, and for a Sub-dataset (as described in~\cite{Alayba2018}) the obtained accuracy was between 0.87 and 0.95.

The sentiment analysis approach in this paper is a combination of two deep neural networks, i.e., a Convolutional Neural Network (CNN) and a Long Short Term Memory (LSTM) network. Kim~\cite{Kim2014} defined CNNs to have convolving filters over each input layer in order to generate the best features. CNNs have shown improvements in computer vision, natural language processing and other tasks. Athiwaratkun and Kang~\cite{Athiwaratkun2015} confirmed that the CNN is a powerful tool to select features in order to improve the prediction accuracy. Gers et al.~\cite{Gers2002} showed the capabilities of LSTMs in learning data series by considering the previous outputs.

This paper first presents some background on deep neural networks and Arabic sentiment classification in Section 2. Section 3 describes the Arabic sentiment datasets we use. Section 4 illustrates the architecture of the proposed merged CNN-LSTMs Arabic sentiment analysis model. The results of the sentiment classification using the model will be presented in Section 5, which  will be compared with other results. Section 6 concludes the study and the experiments,  and outlines the future work.

\section{2.	Background and Related Work}
Deep neural network models have had great success in machine learning, particularly in various tasks of NLP. For example, automatic summarization~\cite{Yousefi-Azar2017}, question answering~\cite{Yih2015}, machine translation~\cite{Auli2013}, words and phrases distributed representations~\cite{Mikolov2013}, sentiment analysis~\cite{Kim2014} and other tasks. Kim~\cite{Kim2014} proposed a deep learning model for sentiment analysis using CNNs with different convolutional filter sizes. Wang et al.~\cite{Wang2016} applied an attention-based LSTMs model for aspect-level sentiment analysis.

% Kalchbrenner et al~\cite{Kalchbrenner2014} introduced a dynamic k-max polling function for a dynamic CNN.

Arabic sentiment analysis has become a research area of interest in recent years. Abdul-Mageed et al.~\cite{Abdul-Mageed2011} studied the effect at sentence level on the subjectivity and sentiment classification for Modern Standard Arabic language (MSA) using an SVM classifier. Shoukry and Rafea~\cite{Shoukry2012} applied SVM and Na\"{\i}ve Bayes at sentence level for sentiment classification using 1000 tweets. Abdulla et al.~\cite{Abdulla2013} compared corpus-based and lexicon-based approaches for sentiment analysis. Abdulla et al.~\cite{Abdulla2014} addressed the challenges of lexicon construction and sentiment analysis. Badaro et al~\cite{Badaro2014} created a large Arabic sentiment lexicon using English-based linking to
the ESWN lexicon and WordNet approach. Duwairi et al.~\cite{Duwairi2014} collected over 300,000 Arabic tweets and labeled over 25,000 tweets using crowdsourcing. Al Sallab et al.~\cite{Al Sallab2015} employed three deep learning methods for Arabic sentiment classification. Ibrahim et al.~\cite{Ibrahim2015} showed sentiment classifications for MSA and the Egyptian dialect using different types of text data such as tweets, product reviews, etc. Dahou et al.~\cite{Dahou2016} reported on the usage of Arabic pre-trained word representation with CNN increased sentiment classification performance. Tartir and Abdul-Nabi~\cite{Tartir2017} concluded that a semantic approach leads to good sentiment classification results even when the dataset size is small. El-Beltagy et al.~\cite{El-Beltagy2018} enhanced the performance of a sentiment classification using a particular set of features.

\section{3.	Datasets}
There is a lack of Arabic sentiment datasets in comparison to English. In this paper, four datasets (where one is a subset of another) will be used in the experiments. Each used only two sentiment classes, i.e., Positive and Negative sentiment.
\subsection{Arabic Health Services Dataset (Main-AHS and Sub-AHS)}

This is our own Arabic sentiment analysis dataset collected from Twitter. It was first presented in~\cite{Alayba2017} and it has two classes (positive and negative). The dataset contains 2026 tweets and it is an unbalanced dataset that has 1398 negative tweets and 628 positive tweets. We call this dataset \textbf{Main-AHS}, and we selected a subset of this dataset, called \textbf{Sub-AHS}, which was introduced in~\cite{Alayba2018}. The \textbf{Sub-AHS} dataset contains 1732 tweets, with 502 positive tweets and 1230 negative tweets.

\subsection{Twitter Data Set (Ar-Twitter)} 
The authors of~\cite{Abdulla2013} have manually built a labeled sentiment analysis dataset from Twitter using a crawler. The dataset contains 2000 tweets with two classes (Positive and Negative) and each class contains 1000 tweets. The dataset covered several topics in Arabic such as politics, communities and arts. There are some tweets in the available online dataset are missing and, hence, the used size of the dataset in our experiments is 975 negative tweets and 1000 positive tweets.

\subsection{Arabic Sentiment Tweets Dataset (ASTD)}
The authors of~\cite{Nabil2015} presented a sentiment analysis dataset from Twitter that contains over 54,000 Arabic tweets. It has four classes (objective, subjective positive, subjective negative, and subjective mixed). However, in this paper only two classes (positive and negative) will be used and the numbers of negative and positive tweets are 1684 and 795 respectively, giving a total of 2479 tweets.

\section{3.	CNN-LSTM Arabic Sentiment Analysis Model}

\begin{figure}
	\centering
	\includegraphics[width=1\textwidth]{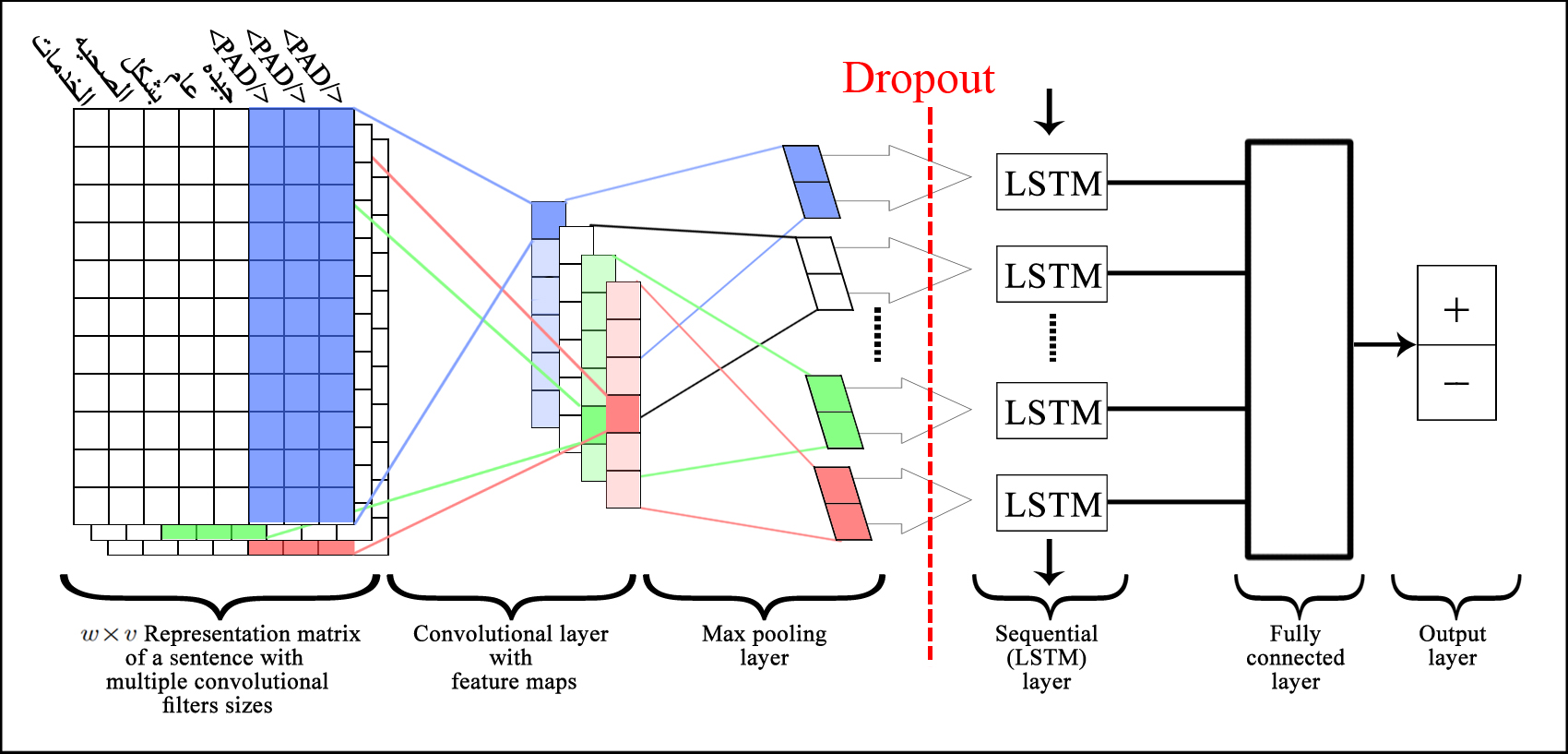}
	\caption{A Combined CNN-LSTM Model architecture for an Arabic example sentence.} \label{Model}
\end{figure}

The fundamental architecture of the proposed model is shown in Figure~\ref{Model} and it outlines the combination of the two neural networks: CNN and LSTM. There are no accurate tools for preprocessing Arabic text, especially non Standard Arabic text like most of the tweets. There are many forms for a single word in Arabic, for example Arabic words are different based on gender, the tenses of the verbs, the speaker voices, etc.~\cite{Bahloul2008}. Table~\ref{table:multipleforms} shows several examples of a single Arabic verb (and it has more other forms), the pronunciation of the word as Buckwalter translation \cite{Smr2016} and the description of the verb's type.

\begin{table}
	\caption{Some examples of multiple forms of a single Arabic verb}
	\label{table:multipleforms}
	\centering
	
	\begin{tabular}{{c} {c} {c}}
		\toprule
		
		\bfseries  {\makecell{Arabic \\ word}}  & \bfseries {\makecell{ Buckwalter \\ Arabic Encoding} } & \bfseries {Word type} 
		\\
		\midrule 
		
		\<فعل> & fEl  & Masculine Verb - past tense for singular\\

		\hline
		\<فعلت> & fElt  & Feminine Verb - past tense for singular\\
		
		\hline
		\<يفعل> & yfEl  & Masculine Verb - present tense for singular\\
		
		\hline
		\<تفعل> & tfEl  & Feminine Verb - present tense for singular\\
		
		\hline
		\<يفعلان> & yfElAn  & Masculine Verb - present tense for dual\\
		
		\hline
		\<تفعلان> & tfElAn  & Feminine Verb - present tense for dual\\
		
		\hline
		\<يفعلون> & yfElwn  & Masculine Verb -  present tense for plural \\
		
		\hline
		\<يفعلن> & yfEln  & Feminine Verb -  present tense for plural \\

		%\midrule
		
		\bottomrule
	\end{tabular}
\end{table}

There will be three different levels of sentiment analysis for each proposed dataset. The reason of using different levels is to try to expand the number of features in short tweets and to deal with many forms of a single word in Arabic. This is an example tweet 
"\<الخدمات الصحيه بشكل عام جيده>"
and the English translation of this tweet is
'Health services are generally good'. The levels are as follows.

\textbf{Character Level (Char-level)}, by converting the sentence into characters instead of words such as [
'\<ا>'
, '\<ل>'
, '\<خ>'
, '\<د>'
, '\<م>'
, '\<ا>'
, '\<ت>'
, '\<ا>'
, '\<ل>'
, '\<ص>'
, '\<ح>'
, '\<ي>'
, '\<ه>'
, '\<ب>'
, '\<ش>'
, '\<ك>'
, '\<ل>'
, '\<ع>'
, '\<ا>'
, '\<م>'
, '\<ج>'
, '\<ي>'
, '\<د>'
, '\<ه>'
]. The (Char-level) for the English example is
['H', 'e', 'a', 'l', 't', 'h', 's', 'e', 'r', 'v', 'i', 'c', 'e', 's', 'a', 'r', 'e', 'g', 'e', 'n', 'e', 'r', 'a', 'l', 'l', 'y',  'g', 'o', 'o', 'd'].
At this level, the number of features is increased, such as in the above example, the number of characters is 24 for the Arabic example, and each letter represents one feature. 

The second level is \textbf{Character \textit{N}-Gram Level (Ch5gram-level)}: where we measure the length of all the words in each dataset and we calculate the average length of words (which is five characters for all the different datasets). Then, we split any word that, has more than the average number into several sub-words. Whereas, any word that consist of the same average number of characters or less will be kept as it is. The average word's length for each dataset is five characters and a 5-gram example is [
'\<الخدم>'
, '\<لخدما>'
, '\<خدمات>'
, '\<الصحي>'
, '\<لصحيه>'
, '\<بشكل>'
, '\<عام>'
, '\<جيده>'
]. The (Ch5gram-level) for the English example is
['Healt', 'ealth', 'servi', 'ervic', 'rvice', 'vices', 'are', 'gener', 'enera', 'neral', 'erall', 'rally', 'good']. This level can be useful in order to deal with many forms of Arabic words, especially for words with more than five letters. Also, the number of the features is expanded in this level too. The third level is \textbf{Word Level (Word-level)}, where the sentence is divided into words using the space as splitter, such as
[
'\<الخدمات>'
, '\<الصحيه>'
, '\<بشكل>'
, '\<عام>'
, '\<جيده>'
]. 

The (Word-level) for the English example is
['Health', 'services', 'are', 'generally', 'good']. This level is the most commonly chosen option in the field of sentiment analysis.

The input data layer is represented as a fixed-dimension matrix of different vector embeddings based on different sentiment analysis levels. Each sentiment analysis level has different tokens, for example, in the Char-level the token is a single character. In the Ch5gram-level, the token is a whole word if the length of the word is five characters or less. Also, the token for any words that has more than five letters is split into five gram character like in the Ch5 Gram-level example from above. In the Word-level, the tokens are based on the words in each tweet. Each token is represented as a fixed-size vector in the input matrix. Then, the multiple convolutional filters slide over the matrix to produce a new feature map and the filters have various different sizes to generate different features. The Max-pooling layer is to calculate the maximum value as a corresponding feature to a specific filter. The output vectors of the Max-pooling layer become inputs to the LSTM networks to measure the long-term dependencies of feature sequences. The output vectors of the LSTMs are concatenated and an activation function is applied to generate the final output: either positive or negative.

\subsection{Input Layer}
This is the first layer in the model and it represents each tweet as a row of vectors. Each vector represents a token based on the the sentiment analysis level used. Each different level has a different token to be embedded, such as in the Char-level, each character in the tweet will be represented into a specific vector with a fixed size of 100. Each word in the tweet, which is one token in the Word-level is embedded into a vector with lengh of 100 and that is the same with each token in the Ch5gram-level. This layer is a matrix of size \textit{w}$\times$\textit{v}, where \textit{v} is the lengh of the vector and \textit{w} is the number of tokens in the tweets. The value of \textit{w} is the maximum length of a tweet. Any tweet that contains less than the maximum number of tokens in the tweet  will be padded with $<$\textit{Pad}$>$ to have the same lengh with the maximum tweet lengh. For instance, the maximum length of tweets with the character level in the Main-AHS dataset is 241 tokens and any tweets that have less than the maximum number will be padded to 241 to get the same length. Each matrix in the Character level in the Main-AHS dataset has the size of 241$ \times $100.

%The alphabet in Arabic has 28 letters, so the embedding vectors size is (28 $ \times $ 100). 

\subsection{Convolutional Layer}
Each input layer contains a sequence of vectors and it is scanned using a fixed size of filter. For example, we used the filter size 3 for Word-level to extract the 3-gram features of words. Also, we used the filter size 20 in the Char-level and the filter size 10 in the Ch5gram-level. The filter strides or shifts only one column and one row over the matrix. Each filter detects multiple features in a tweet using the \textit{ReLU}~\cite{Keras} activation function, in order to represent them in the feature map.

\subsection{Max-Pooling Layer}
After the Convolutional layer, the Max-pooling layer minimizes and down-samples the features in the feature map. The \textit{max} operation or function is the most commonly used technique for this layer and it is used in this experiment. The reason of selecting the highest value is to capture the most important feature and reduce the computation in the advanced layers. Then the dropout technique is applied to reduce overfitting with the dropout value is 0.5.

\subsection{LSTM Layer}
One of the advantages of the LSTMs is the ability of capturing the sequential data by considering the previous data. This layer takes the output vectors from the dropout layer as inputs. This layer has a set number of units or cells and the input of each cell is the output from the dropout layer. The final output of this layer have the same number of units in the network.

\subsection{Fully Connected Layer}
The outputs from LSTMs are merged and combined in one matrix and then passed to a fully connected layer. The array is converted into a single output in the range between 0 and 1 using the fully connected layer, in order to be finally classified using \textit{sigmoid} function \cite{Han1995}.

\section{5.	Experiments and Results}
These experiments aimed to utilize a very deep learning model using a combination of CNN and LSTM. The learning performance of the model will be measured using the accuracy of the classifier~\cite{Manning2008}.

\begin{equation} \label{Accuracy_eq}
Acc = \frac{(TP+TN)}{(TP+TN+FP+FN)} .
\end{equation}	

Here, \textit{TP} is the number of tweets that are positive and predicted correctly as positive, \textit{TN} is the number of tweets that are negative and predicted correctly as negative, \textit{FP} is the number of tweets that are negative but predicted incorrectly as positive, and \textit{FN} is the number of tweets that are positive but predicted incorrectly as negative.

\begin{table}
	\caption{Accuracy comparison of the proposed method with different sentiment levels and other models for the same datasets.}\label{ACC-1}
	\centering
	\begin{tabular}{@{} l *{4}{c} @{}}
		\toprule
		\bfseries  {Sentiment Level  }  & \bfseries {     Main-AHS  } & \bfseries {      Sub-AHS  } & \bfseries {  Ar-Twitter  } & \bfseries {      ASTD}\\
		\midrule 
		Char-level & 0.8941	& 0.9164 & 0.8131 & 0.7419  \\ 
		Ch5gram-level &	0.9163 & \textbf{\underline{0.9568}} & 0.8283	& \underline{0.7762}  \\ 
		Word-level & \textbf{\underline{0.9424}} & 0.9510 & \textbf{\underline{0.8810}} & 0.7641 \\ 
		
		\midrule
		
		\textit{Alayba et al., 2018}~\cite{Alayba2018} & 0.92 & 0.95 &  &  \\
		
		\textit{Dahou et al., 2016}~\cite{Dahou2016} &  &  & 85.01 & \textbf{79.07} \\
		
		\textit{Abdulla et al., 2013}~\cite{Abdulla2013} &  &  & 87.20 &  \\
		
		\bottomrule
	\end{tabular}
\end{table}

\begin{figure}
	\centering
	\includegraphics[width=1\textwidth]{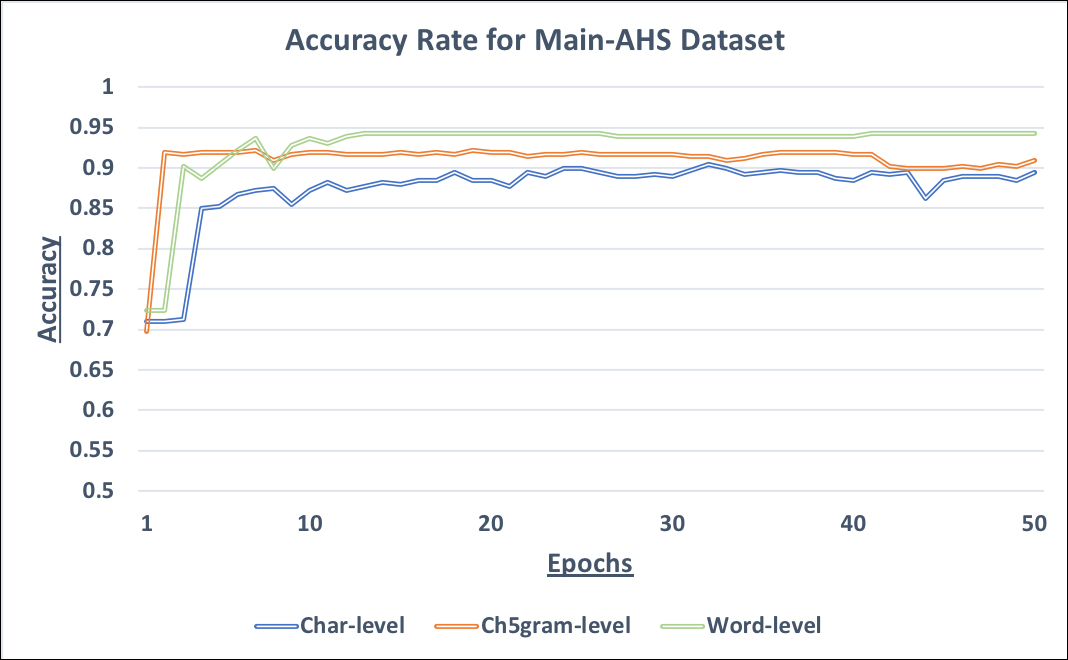}
	\caption{Accuracy on the test set for Main-AHS dataset using different sentiment analysis levels.} \label{AccMain}
\end{figure}

\begin{figure}
	\centering
	\includegraphics[width=1\textwidth]{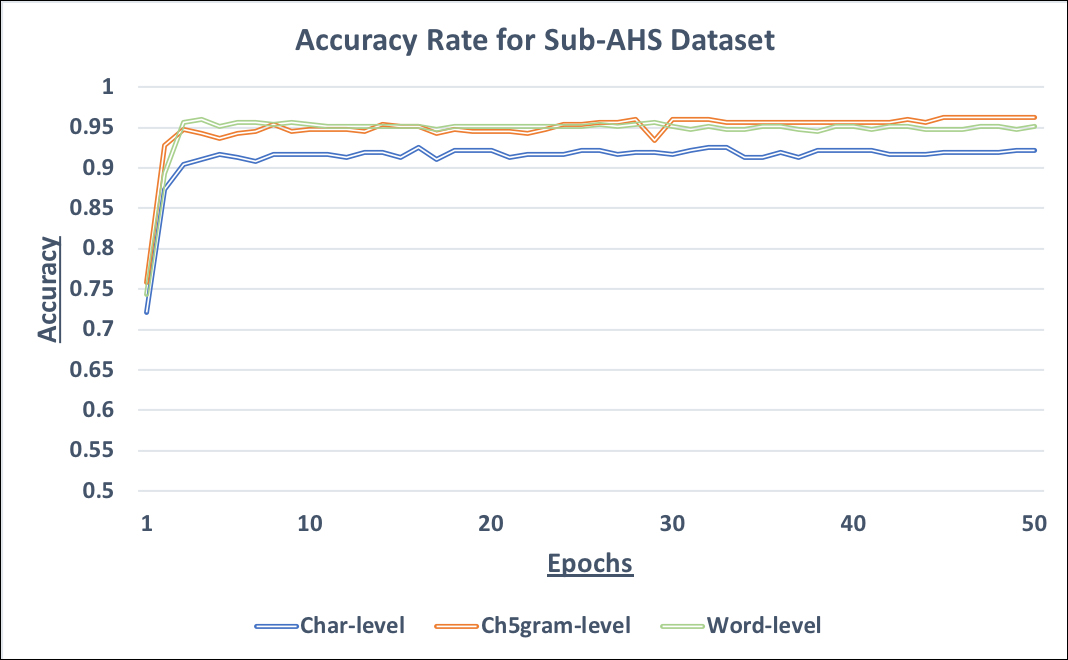}
	\caption{Accuracy on the test set for Sub-AHS dataset using different sentiment analysis levels.} \label{AccSub}
\end{figure}

\begin{figure}
	\centering
	\includegraphics[width=1\textwidth]{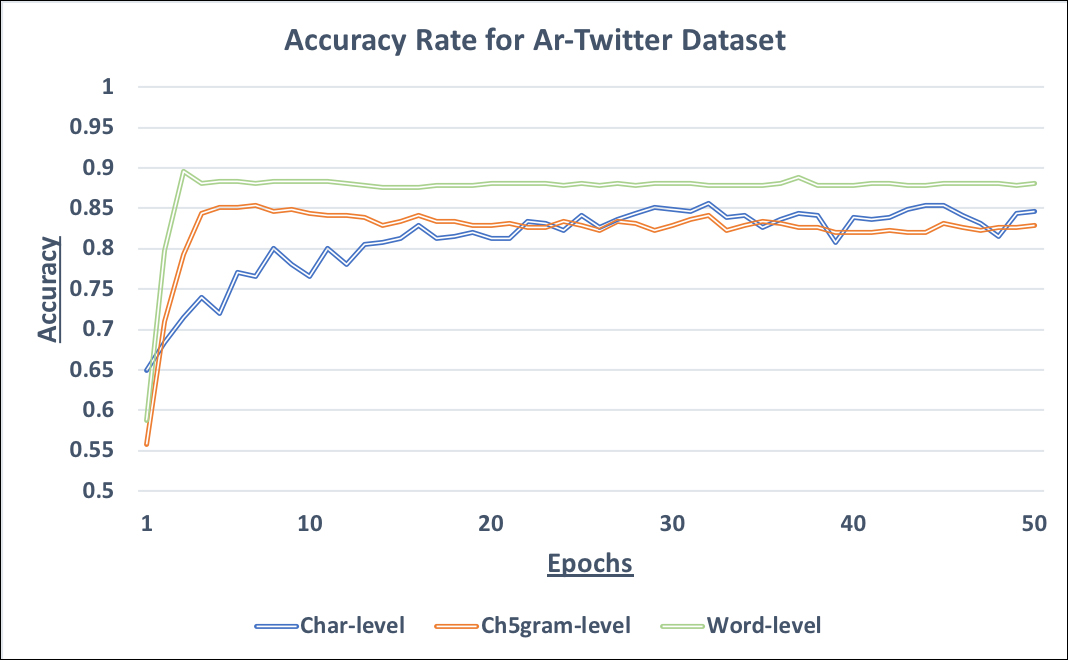}
	\caption{Accuracy on the test set for Ar-Twitter dataset using different sentiment analysis levels.} \label{AccAr-Twitter}
\end{figure}

\begin{figure}
	\centering
	\includegraphics[width=1\textwidth]{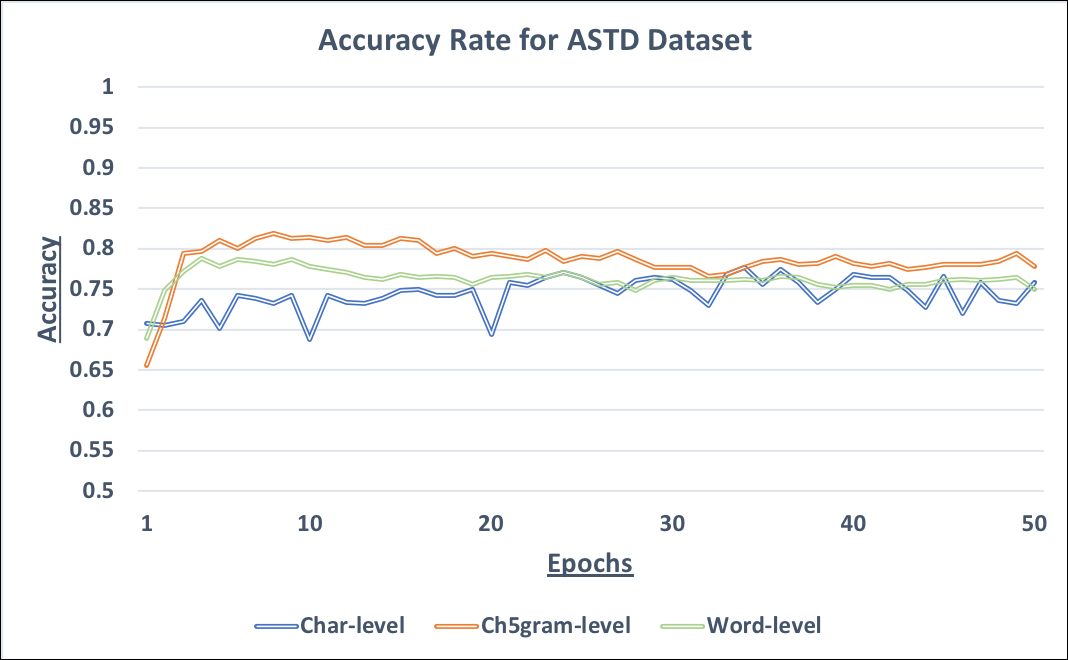}
	\caption{Accuracy on the test set for ASTD dataset using different sentiment analysis levels.} \label{AccASTD}
\end{figure}

All the experiments using different datasets and sentiment analysis levels use the same size of the training and test datasets. The size of the training set is 80\% of the whole dataset, and the test set contains the remaining 20\% of the dataset. The model is trained using the training set and then the test set is used to measure the performance of the model. The number of epochs is 50 for all the experiments. Table~\ref{ACC-1} shows the accuracy results in the 50 epochs for the four datasets using different sentiment levels. The best accuracy results for the three different used levels are identified by underlining the best results. Also, Table~\ref{ACC-1} compares the results of our model with the results published in other papers.
It is clear from Table~\ref{ACC-1} that the proposed model improved the performance of sentiment classification in three datasets: Main-AHS, Sub-AHS, and Ar-Twitter, but it is lower than~\cite{Dahou2016} for the ASTD dataset model (by only a small margin).
Figures~\ref{AccMain},~\ref{AccSub},~\ref{AccAr-Twitter}, and~\ref{AccASTD} illustrate the accuracies on different datasets over 50 epochs. Each line represents different sentiment analysis level. Char-level generally has the lowest accuracy results in the different datasets compared with the other levels, but for Ar-Twitter, it is better than the accuracy obtained on the Ch5gram-level after 23 epochs. Word-level achieves the best accuracy results for the Main-AHS and Ar-Twitter datasets and it has similar results with Ch5gram-level for the Sub-AHS.

\section{6.	Conclusions and Future Work}

This paper investigated the benefits of combining CNNs and LSTMs networks in an Arabic sentiment classification task. It also explored the effectiveness of using different levels of sentiment analysis because of the complexities of morphology and orthography in Arabic. We used character level to increase the number of features for each tweet, as we are dealing with short messages, which was not an ideal option for our model. However, using Word-level and Ch5gram-level have shown better sentiment classification results.

This approach has improved the sentiment classification accuracy for our Arabic Health Services (AHS) dataset to reach 0.9424 for the Main-AHS dataset, and 0.9568 for the Sub-AHS dateset, compared to our previous results in~\cite{Alayba2018} which were 0.92 for the Main-AHS dataset and 0.95 for the Sub-AHS dateset.
Future work will use some pre-trained word representation models, such as word2vec~\cite{Mikolov2013}, GloVe~\cite{Pennington2014}, and Fasttext~\cite{Bojanowski2017} for the embedding layer.

%
% the environments 'definition', 'lemma', 'proposition', 'corollary',
% 'remark', and 'example' are defined in the LLNCS documentclass as well.
%


\begin{thebibliography}{8}
	
	
 
%%% 1 %%% 
\bibitem{Liu2012}
Liu, B.: Sentiment Analysis and Opinion Mining. Morgan \& Claypool, (2012).


%%% 2 %%% 
\bibitem{Balaji2017}
Balaji, P., Nagaraju, O., Haritha, D.: Levels of sentiment analysis and its challenges: A literature review. In: International Conference on Big Data Analytics and Computational Intelligence (ICBDAC) 2017, vol. 6, pp. 436--439.
Chirala, India, (2017).


%%% 3 %%% 
\bibitem{Feldman2013}
Feldman, R.: Techniques and Applications for Sentiment Analysis. Commun. ACM 56(4), 82--89 (2013).


%%% 4 %%% 
\bibitem{Lakomkin2017}
Lakomkin, E., Bothe, C., Wermter, S.: GradAscent at EmoInt-2017: Character and Word Level Recurrent Neural Network Models for Tweet Emotion Intensity Detection. In: Editor,
F., Editor, S. (eds.) Proceedings of the 8th Workshop on Computational Approaches to Subjectivity, Sentiment and Social Media Analysis 2017, pp. 169--174.
ACL, Copenhagen, Denmark (2017). 


%%% 5 %%% 
\bibitem{Joshi2016}
Joshi, A., Prabhu, A., Shrivastava, M., Varma, V.: Towards Sub-Word Level Compositions for Sentiment Analysis of Hindi-English Code Mixed Text. Proceedings of COLING 2016, the 26th International Conference on Computational Linguistics: Technical Papers 2016, pp. 2482--2491.
The COLING 2016 Organizing Committee, Osaka, Japan (2016). 	


%%% 6 %%% 
\bibitem{Kim2014}
Kim, Y.: Convolutional Neural Networks for Sentence Classification. Proceedings of the 2014 Conference on Empirical Methods in Natural Language Processing (EMNLP), pp. 1746--1751.
ACL, Doha, Qatar (2014). 


%%% 7 %%% 
\bibitem{Yin2015}
Yin, W., Sch{\"u}tze, H.: Multichannel Variable-Size Convolution for Sentence Classification. Proceedings of the Nineteenth Conference on Computational Natural Language Learning, pp. 204--214. ACL, Beijing, China (2015). 


%%% 8 %%% 
\bibitem{Shin2017}
Shin, B., Lee, T., Choi, J. D.: Lexicon Integrated CNN Models with Attention for Sentiment Analysis. Proceedings of the 8th Workshop on Computational Approaches to Subjectivity, Sentiment and Social Media Analysis, pp. 149--158. ACL, Copenhagen, Denmark (2017). 

%%% 9 %%% 
\bibitem{Alayba2017}
Alayba, A. M., Palade, V., England, M., Iqbal, R.: Arabic Language Sentiment Analysis on Health Services. In: 2017 1st International Workshop on Arabic Script Analysis and Recognition (ASAR), pp. 114--118, IEEE, Nancy, France (2017). 



%%% 10 %%% 
\bibitem{Alayba2018}
Alayba, A. M., Palade, V., England, M., Iqbal, R.: Improving Sentiment Analysis in Arabic Using Word Representation. In: 2018 2nd International Workshop on Arabic and Derived Script Analysis and Recognition (ASAR), pp. 13--18, IEEE, London, UK (2018).






%%% 11 %%% 
\bibitem{Athiwaratkun2015}
Athiwaratkun, B., Kang, K.: Feature Representation in Convolutional Neural
Networks. arXiv preprint arXiv:1507.02313, (2015).



%%% 12 %%% 
\bibitem{Gers2002}
Gers F.A., Eck D., Schmidhuber J.: Applying LSTM to Time Series Predictable Through Time-Window Approaches. In: Editor,
Tagliaferri R., Marinaro M. (eds.) Neural Nets WIRN Vietri-01. Perspectives in Neural Computing 2002., vol. 9999, pp. 193--200. Springer, London (2002). 



%%% 13 %%% 
\bibitem{Yousefi-Azar2017}
Yousefi-Azar, M., Hamey, L.: Text Summarization Using Unsupervised Deep Learning. Expert Systems with Applications 68, 93--105 (2017).


%%% 14 %%% 
\bibitem{Yih2015}
Yih, S. W., He, X., Meek, C.: Semantic Parsing for Single-Relation Question Answering. In: Proceedings of the 52nd Annual Meeting of the Association for Computational Linguistics, (vol. 2: Short Papers), pp. 643--648. ACL, Baltimore, Maryland, USA (2014). 



%%% 15 %%% 
\bibitem{Auli2013}
Auli, M. W., Galley, M., Quirk, C., Zweig, G.: Joint Language and Translation Modeling with Recurrent Neural Networks. In: Proceedings of the 2013 Conference on Empirical Methods in Natural Language Processing, pp. 1044--1054. ACL, Seattle, Washington, USA (2013). 



%%% 16 %%% 
\bibitem{Mikolov2013}
Mikolov, T., Sutskever, I., Chen, K., Corrado, G., Dean, J.: Distributed Representations of Words and Phrases and Their Compositionality. In: Proceedings of the 26th International Conference on Neural Information Processing Systems NIPS'2013, vol. 2, pp. 3111--3119. Curran Associates Inc., Lake Tahoe, Nevada, USA (2013). 



%%% 17 %%% 
%\bibitem{Kalchbrenner2014}
%Kalchbrenner, N., Grefenstette, E., Blunsom, P.: A convolutional neural network for modelling sentences. In: Proceedings of the 52nd Annual Meeting of the Association for Computational Linguistics ACL 2014, vol. 1, pp. 655--665. ACL, Baltimore, MD, USA (2014). 



%%% 18 %%% 
\bibitem{Wang2016}
Wang, Y., Huang, M., Zhao, L., Zhu, X.: Attention-based LSTM for Aspect-level Sentiment Classification. In: Proceedings of the 2016 Conference on Empirical Methods in Natural Language Processing 2016, pp. 606--615. ACL, Austin, Texas (2016). 




%%% 19 %%% 
\bibitem{Abdul-Mageed2011}
Abdul-Mageed, M., Diab, M. T., Korayem, M.: Subjectivity and Sentiment Analysis of Modern Standard Arabic. In: Proceedings of the 49th Annual Meeting of the Association for Computational Linguistics: Human Language Technologies HLT 2011: Short Papers - Volume 2, pp. 587--591. ACL, Stroudsburg, PA, USA (2011). 




%%% 20 %%% 
\bibitem{Shoukry2012}
Shoukry, A., Rafea, A.: Sentence-Level Arabic Sentiment Analysis. In: 2012 International Conference on Collaboration Technologies and Systems (CTS), pp. 546--550, IEEE, Denver, CO, USA (2012). 


%%% 21 %%% 
\bibitem{Abdulla2013}
Abdulla, N. A., Ahmed, N. A., Shehab, M. A., Al-Ayyoub, M.: Arabic sentiment analysis: Lexicon-based and corpus-based. In: 2013 IEEE Jordan Conference on Applied Electrical Engineering and Computing Technologies (AEECT), pp. 1--6, IEEE, Amman, Jordan (2013). 


%%% 22 %%% 
\bibitem{Abdulla2014}
Abdulla, N., Majdalawi, R., Mohammed, S., Al-Ayyoub, M., Al-Kabi, M.: Automatic Lexicon Construction for Arabic Sentiment Analysis. In: 2014 International Conference on Future Internet of Things and Cloud, pp. 547--552, IEEE, Barcelona, Spain (2014). 



%%% 23 %%% 
\bibitem{Badaro2014}
Badaro, G., Baly, R., Hajj, H., Habash, N., El-Hajj, W.: A Large Scale Arabic Sentiment Lexicon for Arabic Opinion Mining. In: Proceedings of the EMNLP 2014 Workshop on Arabic Natural Language Processing (ANLP), pp. 165--173, ACL, Doha, Qatar (2014). 



%%% 24 %%% 
\bibitem{Duwairi2014}
Duwairi, R. M., Marji, R., Sha'ban, N., Rushaidat, S.: Sentiment Analysis in Arabic tweets. In: 2014 5th International Conference on Information and Communication Systems (ICICS), pp. 1--6, IEEE, Irbid, Jordan (2014). 




%%% 25 %%% 
\bibitem{Al Sallab2015}
Al Sallab, A., Hajj, H., Badaro, G., Baly, B., El Haj, W., Shaban, K. B.: Deep Learning Models for Sentiment Analysis in Arabic. In: Proceedings of the Second Workshop on Arabic Natural Language Processing, pp. 9--17, ACL, Beijing, China (2015). 




%%% 26 %%% 
\bibitem{Ibrahim2015}
Ibrahim, H. F., Abdou, S. M., Gheith, M.: Sentiment Analysis for Modern Standard Arabic and Colloquial. International Journal on Natural Language Computing (IJNLC) 4(2), 95--109 (2015).



%%% 27 %%% 
\bibitem{Dahou2016}
Dahou, A., Xiong, S., Zhou, J., Haddoud, M. H., Duan, P.: Word Embeddings and Convolutional Neural Network for Arabic Sentiment Classification. In: Proceedings of COLING 2016, the 26th International Conference on Computational Linguistics: Technical Papers, pp. 2418--2427, The COLING 2016 Organizing Committee, Osaka, Japan (2016). 



%%% 28 %%% 
\bibitem{Tartir2017}
Tartir, S., Abdul-Nabi, I.: ASemantic Sentiment Analysis in Arabic Social Media. Journal of King Saud University - Computer and Information Sciences 29(2), 229--233 (2017).



%%% 29 %%% 
\bibitem{El-Beltagy2018}
El-Beltagy S.R., Khalil T., Halaby A., Hammad M.: Combining Lexical Features and a Supervised Learning Approach for Arabic Sentiment Analysis. In: Gelbukh A. (eds) Computational Linguistics and Intelligent Text Processing. CICLing 2016. Lecture Notes in Computer Science, vol 9624. Springer, Cham (2018).


%%% 30 %%% 
\bibitem{Nabil2015}
Nabil, M., Aly, M., Atiya, A.: ASTD: Arabic Sentiment Tweets Dataset. In: Proceedings of the 2015 Conference on Empirical Methods in Natural Language Processing, pp. 2515--2519, ACL, Lisbon, Portugal (2015).


%%% 32 %%% 
\bibitem{Smr2016}
Smr{\v{z}}, O., Encode Arabic Online Interface, \url{http://quest.ms.mff.cuni.cz/cgi-bin/encode/index.fcgi}. Last accessed 18
June 2018.




%%% 31 %%% 
\bibitem{Bahloul2008}
Bahloul, M.: Structure and Function of the Arabic Verb. Routledge,
London (2008).


%%% 32 %%% 
\bibitem{Keras}
Keras, \url{https://keras.io}. Last accessed 15
Apr 2018.


%%% 33 %%% 
\bibitem{Han1995}
Han, J. and M., Claudio: The Influence of the Sigmoid Function Parameters on the Speed of Backpropagation Learning. In: Proceedings of the International Workshop on Artificial Neural Networks: From Natural to Artificial Neural Computation, pp. 195--201, Springer-Verlag, London, UK (1995). 




%%% 34 %%% 
\bibitem{Manning2008}
Manning, C., Raghavan, P., Sch\"{u}tze, H.: Introduction to Information Retrieval.  1st edn. Cambridge University Press,
New York, NY, USA (2008).


%%% 35 %%% 
\bibitem{Pennington2014}
Pennington, J., Socher, R., Manning, C. D.: GloVe: Global Vectors for Word Representation. In: Empirical Methods in Natural Language Processing (EMNLP), pp. 1532--1543, ACL, Doha, Qatar (2014). 


%%% 36 %%% 
\bibitem{Bojanowski2017}
Bojanowski, P., Grave, E., Joulin, A. Mikolov, T.: Enriching Word Vectors with Subword Information. Transactions of the Association for Computational Linguistics 5, 135--146 (2017).

\end{thebibliography}
%

\end{document}